\documentclass{bmvc2k}

\title{Robustifying the Multi-Scale Representation of Neural Radiance Fields}

\usepackage{enumitem}
\usepackage{amsmath,amssymb}
\usepackage{gensymb}
\addauthor{Nishant Jain}{njain@cs.iitr.ac.in}{1}
\addauthor{Suryansh Kumar}{sukumar@vision.ee.ethz.ch}{2$\dagger$}
\addauthor{Luc Van Gool}{vangool@vision.ee.ethz.ch}{2,3}

\addinstitution{
 Indian Institute of Technology \\
 Roorkee, India
}
\addinstitution{
 ETH Z\"urich \\
 Switzerland
}
\addinstitution{
 KU Leuven \\
 Belgium
}

\pdfoutput=1

\runninghead{Jain, Kumar, Gool}{Robust Multi-Scale Neural Radiance Fields}

\begin{document}

\maketitle

\begin{abstract}
Neural Radiance Fields (NeRF) recently emerged as a new paradigm for object representation from multi-view (MV) images. Yet, it cannot handle multi-scale (MS) images and camera pose estimation errors, which generally is the case with multi-view images captured from a day-to-day commodity camera. Although recently proposed Mip-NeRF could handle multi-scale imaging problems with NeRF, it cannot handle camera pose estimation error. On the other hand, the newly proposed BARF can solve the camera pose problem with NeRF but fails if the images are multi-scale in nature. This paper presents a robust multi-scale neural radiance fields representation approach to simultaneously overcome both real-world imaging issues. Our method handles multi-scale imaging effects and camera-pose estimation problems with NeRF-inspired approaches by leveraging the fundamentals of scene rigidity. To reduce unpleasant aliasing artifacts due to multi-scale images in the ray space, we leverage Mip-NeRF multi-scale representation. For joint estimation of robust camera pose, we propose graph-neural network-based multiple motion averaging in the neural volume rendering framework. We demonstrate, with examples, that for an accurate neural representation of an object from day-to-day acquired multi-view images, it is crucial to have precise camera-pose estimates. Without considering robustness measures in the camera pose estimation, modeling for multi-scale aliasing artifacts via conical frustum can be counterproductive. We present extensive experiments on the benchmark datasets to demonstrate that our approach provides better results than the recent NeRF-inspired approaches for such realistic settings.
\end{abstract}

\section{Introduction}\label{sec:intro}
NeRF has emerged as a popular method of choice for object representation from its multi-view (MV) images \cite{mildenhall2020nerf}. This new 3D representation has shown promising results on several computer vision,  graphics and robotics problems \cite{yu2021pixelnerf,sucar2021imap,zhang2021nerfactor,liu2020neural,martel2021acorn, kaya2022neural, lee2022uncertainty}. Yet, it has some inherent challenges in handling day-to-day captured multi-view images. For instance, NeRF shows observable artifacts on multiple scale images \cite{barron2021mipnerf}, and its performance degrades even with subtle inaccuracies in camera pose estimates \cite{lin2021barf}. Now, most of us might have experienced that with everyday commodity cameras, it is challenging, if not impossible, to acquire an object's MV images at the same scale and recover correct camera poses using them simultaneously. On the other hand, popular off-the-self pose solvers such as COLMAP \cite{schoenberger2016sfm} have limitations in providing accurate camera poses \cite{chatterjee2017robust, kataria2020improving}, which inherently limits the broader application of NeRF. Such limitations with NeRF were easily noticeable, leading to a few recent follow-ups addressing those limitations with NeRF, yet independently.

\begin{figure*}[t]
\centering
\subfigure[\label{fig:first_page_teaser}  \textbf{Left}: Result Illustration  ]{\includegraphics[width=0.48\linewidth]{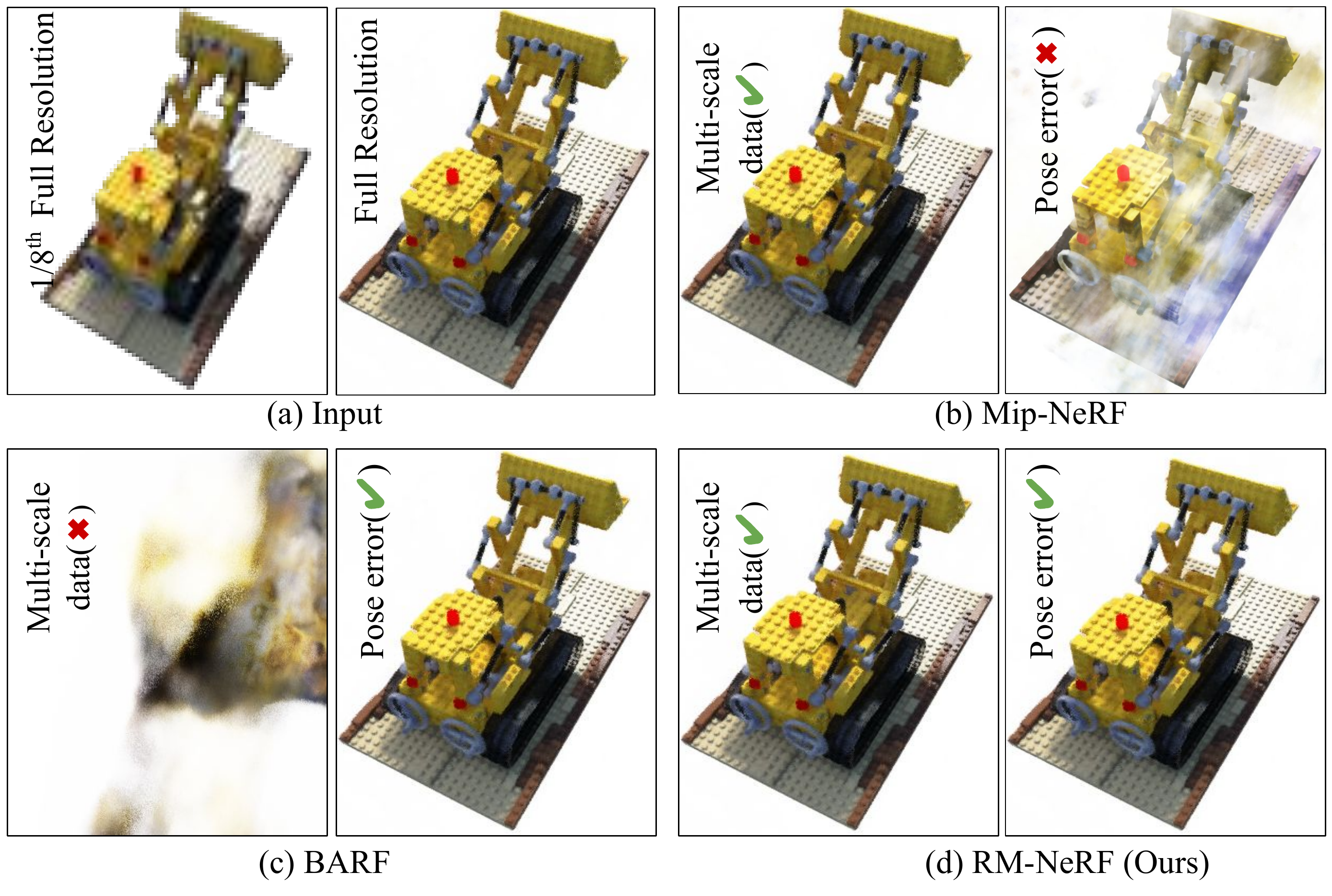}}
~~~\subfigure[\label{fig:mip_cone_illustration}  \textbf{Right}: Intuition on camera pose. ]{\includegraphics[width=0.48\linewidth]{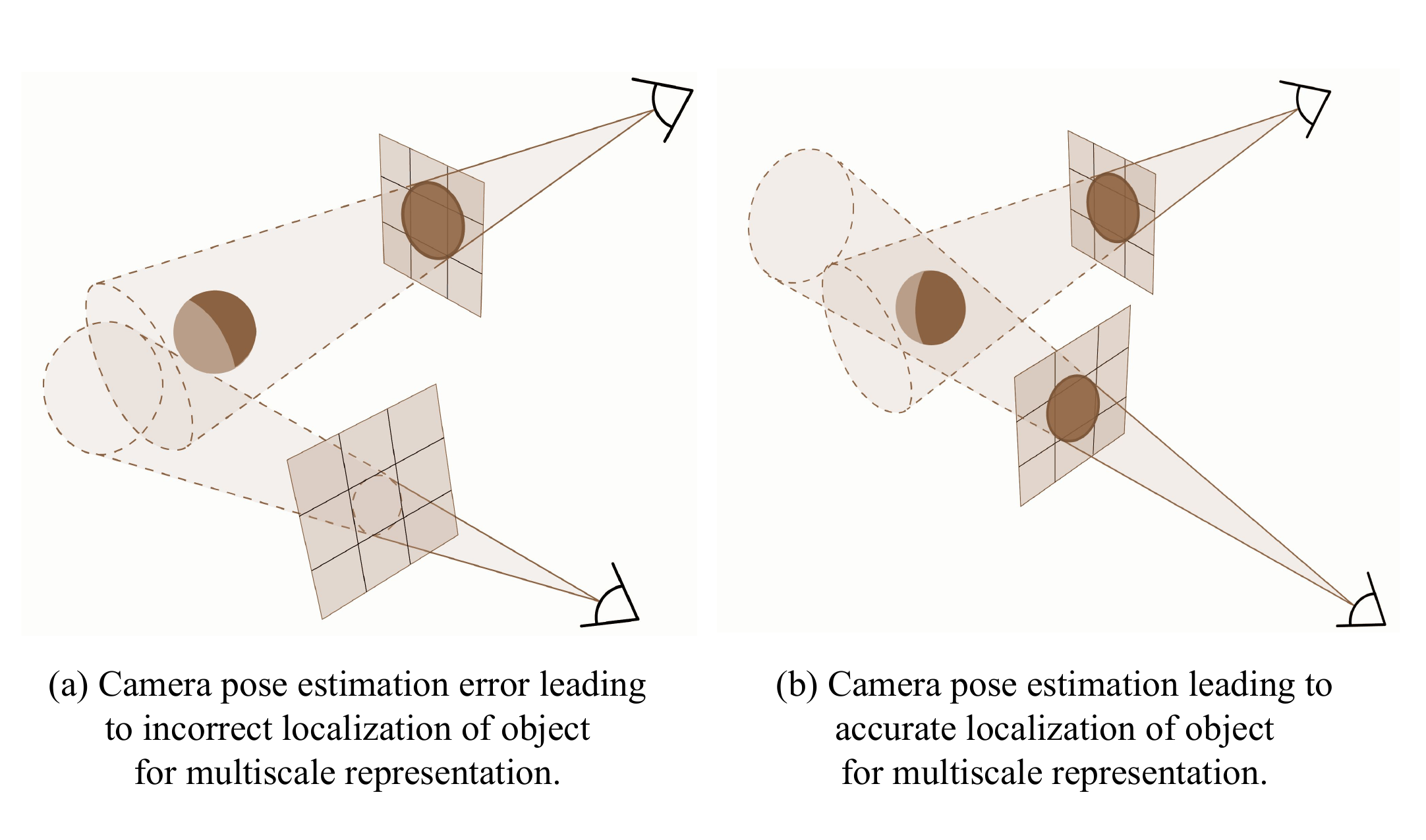}}
\caption{\footnotesize \textbf{Left}: (a) Multi-scaled, multi-view images of Lego example with camera pose error is fed to the NeRF inspired methods \cite{mildenhall2020nerf, barron2021mipnerf, lin2021barf}. (b) \cite{barron2021mipnerf} can handle multi-scale imaging effect but fails if camera pose error also persists. (c) \cite{lin2021barf} can handle the camera pose error for same scale images but fails for multi-scale input images. (d) Our approach works for both cases. \textbf{Right}: A visual illustration: (a) Error in the camera pose estimation can lead to incorrect cones casting in the volume space leading to misguided localization of the object for proper modeling. (b) Correct camera pose certify the proper modeling of the object volume for each sampled canonical frustum.}
\label{fig:examples_fig_page_1}
\end{figure*}

Concretely, the recently proposed BARF \cite{lin2021barf}, and NeRF-- \cite{wang2021nerf} method can overcome the requirement of correct camera pose for NeRF, assuming that images are captured at equidistant from an object. On a separate line of research---inspired by anti-aliasing techniques in computer graphics rendering---the newly proposed Mip-NeRF \cite{barron2021mipnerf} solves the multi-scale problem with NeRF by leveraging the mipmapping approach to rendering. Yet, it assumes ground-truth camera poses are known or estimated well via COLMAP. As is known, ground-truth pose estimation is a challenging task, and a popular framework such as COLMAP  has its challenges in recovering robust camera pose from real-world images \cite{kataria2020improving, chatterjee2017robust}. In both of these groups of independent research, as mentioned above, there exists a gap, \textit{i.e.}, BARF and similar methods can handle the camera pose problem but cannot handle multi-scale image issue, whereas, Mip-NeRF can handle the multi-scale problem but assumes the correct camera pose. In this paper, we propose a simple and effective approach that can fill this gap by utilizing the fundamentals of a rigid scene. Our method jointly addresses the multi-scale problem, and the robust camera poses estimation requirements with neural volume rendering (see Fig.\ref{fig:first_page_teaser}). Furthermore, our approach comprehensively solves NeRF problems and eliminates third-party dependencies for pose estimation, hence a self-contained approach.


Meanwhile, we would like to show that Mip-NeRF multi-scale modeling in the ray space could break down if the camera pose error is not considered. As shown in Fig.\ref{fig:mip_cone_illustration}, the correct intersection of the conical frustum that can localize the object is possible only if both the camera poses are correctly known. One trivial way to handle this is to jointly optimize for object representation and camera pose as done in BARF \cite{lin2021barf} and NeRF-- \cite{wang2021nerf} but using Mip-NeRF \cite{barron2021mipnerf} formulation. However, the bundle-adjustment (BA) based joint optimization is complex, sub-optimal, requires good initialization, and can handle only certain types of noise and outlier distribution \cite{chatterjee2017robust, purkait2020neurora}. And therefore, conditioning the multi-scale rendering representation based on BA-type complex optimization can complicate the overall approach, hence not an encouraging take on the problem.


To solve the challenges mentioned above, we resort to fundamentals of scene rigidity \cite{govindu2006robustness}. If the scene is rigid, we can estimate the camera motion robustly without having explicit information about the object's 3D position. Accordingly, our proposed method initially disentangles the camera pose estimation from neural volume rendering to recover a good pose for joint optimization of the proposed loss function. Our approach uses graph-neural network-based multiple motion averaging with multi-scale feature modeling for the robust camera pose to solve the problem. We evaluated our method's performance on the widely used benchmark dataset \cite{mildenhall2020nerf,barron2021mipnerf}, which clearly shows superior results to the competing baseline methods. In this paper, we make the following contributions.

\smallskip
\noindent
\textbf{Contributions}
\begin{itemize}[leftmargin=*,topsep=0pt, noitemsep]
    \item We propose a method that jointly solves camera pose and multi-scale object representation for day-to-day captured multi-view images using NeRF based representation.
    \item Our method uses scene rigidity fundamentals to jointly optimize camera pose and rendering loss function. To this end, our approach leverage multi-scale representation \cite{barron2021mipnerf} and introduces graph neural network-based multiple motion averaging to learn the noisy camera motion estimates from the images. As a result, our approach helps in better estimation of the network's model parameters for the multi-scale scene or object representation.
    \item The proposed method achieves better camera pose estimates and novel view rendering results than the existing NeRF-based baseline approaches when tested on the standard benchmark sequence \cite{mildenhall2020nerf} and other popular real-world sequences \cite{Knapitsch2017}.
\end{itemize}

\section{Background and Preliminaries}
Recently, implicit neural representation for object or scene inspired by NeRF \cite{mildenhall2020nerf} has gained significant attention in the computer vision and graphics community with many extensions \cite{dellaert2020neural}. Consequently, discussing all the NeRF-related methods is beyond the scope of the paper, and readers may refer to  \cite{dellaert2020neural,tewari2021advances} for a quick reference. Nevertheless, to keep the discussion concise, we discuss the papers directly relevant to our proposed approach.

\subsection{Closely Related Work}
Lately, NeRF has become a popular method of choice for representing a rigid scene as a continuous volumetric field parametrized by a multi-layer perceptron (MLP) \cite{mildenhall2020nerf}. Assuming a {calibrated} setting with {well-posed} input images, NeRF for each pixel sample points along rays that are traced from the camera's center of projection. Further, these sampled points are transformed using positional encoding to represent each point in a high-dimensional feature vector before being fed to an MLP for density and color estimation for novel view synthesis.

\smallskip
\noindent
\textbf{\textit{(i)} Multiscale NeRF.} Barron \textit{et al.} \cite{barron2021mipnerf} introduced Mip-NeRF to overcome the limitation with NeRF in rendering multi-resolution images, \textit{i.e.}, multi-view images that are taken at a different distance from the object. Instead of sampling points along the rays traced from the camera center of projection, Mip-NeRF queries samples along a conical frustum interval region approximated using 3D Gaussian to render the corresponding pixel. As alluded to above, acquisition of images at a perfect scale is unrealistic using day-to-day cameras, and therefore, Mip-NeRF broadens the scope of neural volume rendering approaches to commonly acquired multi-view and multi-scale image acquisition setup. Yet, Mip-NeRF assumption on the availability of ground-truth camera pose parameters is rather unrealistic and could substantially constrain its broader usage.

\smallskip
\noindent
\textbf{\textit{(ii)} Uncalibrated NeRF.} Not long ago, few methods have appeared to solve both for camera pose and object 3D representation extending the neural radiance fields formulation. For instance, BARF \cite{lin2021barf} leverages photometric BA to jointly register the camera poses and recover object representation. On the other hand, NeRF--\cite{wang2021nerf} solves for both intrinsic and extrinsic camera calibration while training NeRF model. Nonetheless, these extensions of NeRF works well only for the same scale images; hence its usage is limited to a synthetic multi-view dome or hemispherical setup. Other related work includes iNeRF \cite{yen2020inerf} that solves camera poses given a well-trained NeRF model.

\subsection{Camera Pose Estimation}
Widely used approaches to camera pose estimation from multi-view images are based on filtering of image key-points and incrementally solve pose \cite{agarwal2011building} or use global BA \cite{Triggs:1999:BAM:646271.685629} that generally has five-point \cite{nister2004efficient} or eight-point algorithm \cite{hartley1997defense} at the back-end. Yet, we know that such methods can provide sub-optimal solutions and may not robustly handle outliers inherent to the unstructured set of images. To address such an intrinsic challenge with pose estimation, Govindu \cite{govindu2001combining} initiated and later authored/co-authored a series of robust multiple rotation averaging (MRA) approaches \cite{govindu2016motion, chatterjee2017robust}. The benefit of using MRA is that it uses multiple estimates of noisy relative motion to solve absolute camera pose based on view-graph representation and rotation group structure \cite{govindu2006robustness} \textit{i.e.}, $SO(3)$. Contrary to the conventional robust rotation averaging approaches \cite{chatterjee2017robust, aftab2014generalized, hartley2011l1, arrigoni2018robust}, in this work, we adhere to recent graph neural network-based approaches \cite{Yang_2021_CVPR, gilmer2017neural, purkait2020neurora,li2021pogo} for robust camera pose estimation.

\section{Proposed Approach}\label{sec:mipnerf}
This paper introduces an approach that unifies two independent research fields in geometric computer vision and volume rendering for scene representation. Our method exploits the multi-scale model of Mip-NeRF and composes it with graph-neural network-based robust motion averaging in a joint optimization cost function. We begin our discussion with multiscale representation for NeRF followed by robust multiple motion averaging.

\smallskip
\noindent
\textbf{\textit{(i)} Multiscale Representation for NeRF.} By leveraging pre-filtering \cite{amanatides1984ray} techniques in rendering  \textit{i.e.}, tracing a cone instead of ray, Mip-NeRF \cite{barron2021mipnerf} learns the scene representation by training a single neural network, which can be queried at arbitrary scales. Further, contrary to NeRF, which uses point-based sampling along each pixel ray to form their positional encoding (PE) feature vector, Mip-NeRF uses the volume of each conical frustum along the cone to model the integrated positional encoding (IPE) features.
The positional encoding $\gamma(\mathbf{x})$ (as defined in NeRF \cite{mildenhall2020nerf}) of all the point within the conical frustum is formulated as:
\begin{equation}
\label{eqn:ipe}
     \gamma^{*}(\textbf{o},\textbf{d},\textit{$\dot{r}$},\textit{$t_{0}$},\textit{$t_{1}$}) = \frac{\int_{}{}\gamma(\textbf{x})\textbf{F}(\textbf{x},\textbf{o},\textbf{d},\textit{$\dot{r}$},\textit{$t_{0}$},\textit{$t_{1}$})\textit{d}\textbf{x}}{\int_{}{}\textbf{F}(\textbf{x},\textbf{o},\textbf{d},\textit{$\dot{r}$},\textit{$t_{0}$},\textit{$t_{1}$})\textit{d}\textbf{x}}
\end{equation}
where $\mathbf{F}$ is an indicator function regarding whether a point lies inside the frustum in the given range [\textit{$t_{0}$}, \textit{$t_{1}$}].
Nevertheless, Eq.\eqref{eqn:ipe} is computationally intractable with no closed form solution and therefore, it is approximated using multivariate Gaussian which provides ``integrated positional encoding'' (IPE) feature\cite{barron2021mipnerf}\footnote{For more details and derivations, kindly refer \cite{barron2021mipnerf}}.

\smallskip
\noindent
\textbf{\textit{(ii)} Scene Rigidity and Multiple Motion Averaging.} 
Assume a pin-hole camera model with known intrinsic calibration matrix $\mathbf{K} \in \mathbb{R}^{3 \times 3}$ with $\mathbf{R} \in SO(3), \mathbf{t} \in \mathbb{R}^{3 \times 1}$ as the rotation and translation w.r.t reference frame.  We can relate $i^{th}$ image pixel $x = [u_i, v_i, 1]^{T}$ to its corresponding 3D point $\mathbf{x} = [{x}_i, {y}_i, {z}_i]^{T}$ as follows:
\begin{equation}\label{eq:img_relation}
[u_i, v_i, 1]^{T} = \mathbf{K}[\mathbf{R}~|~ \mathbf{t}]~[{x}_i, {y}_i, {z}_i, 1]^{T}
\end{equation}

Eq.\eqref{eq:img_relation} indicate a non-linear interaction between 3D scene point and camera motion. 
Yet, the classical epipolar geometry model suggests that if the scene is rigid $x'^{T}\mathbf{E}~x = 0$ must hold \cite{hartleymultiple}, where $x'$ is the image correspondence of $x$ in the next image frame. It is well-studied that $\mathbf{E}$ can be decomposed into $\mathbf{R}~\text{and}~\mathbf{t}$ such that $\mathbf{E} = [\mathbf{t}]_{\times} \mathbf{R}$, where $\mathbf{E} \in \mathbb{R}^{3 \times 3}$ is the essential matrix and $[\mathbf{t}]_{\times} \in \mathbb{R}^{3 \times 3}$ is the skew-symmetric matrix representation of the translation vector \cite{hartleymultiple}.  And therefore, we can estimate rigid motion without making use of any actual 3D observation. Nonetheless, rigid motion solution based on epipolar algebraic relation is not robust to outliers and may provide unreliable results with more multi-view images \cite{chatterjee2017robust}. So to estimate robust camera motion independent of 3D scene point in a computationally efficient way led to the success of robust motion averaging approaches in geometric computer vision \cite{govindu2006robustness, aftab2014generalized, chatterjee2017robust}. 
Further, given rotations, solving translations generally becomes a linear problem \cite{chatterjee2017robust}. Consequently, solution to motion averaging reduces to rotation averaging problem. 

\subsection{Formulation, Loss Function and Optimization}
\begin{figure*}
    \centering
    \includegraphics[scale=0.45]{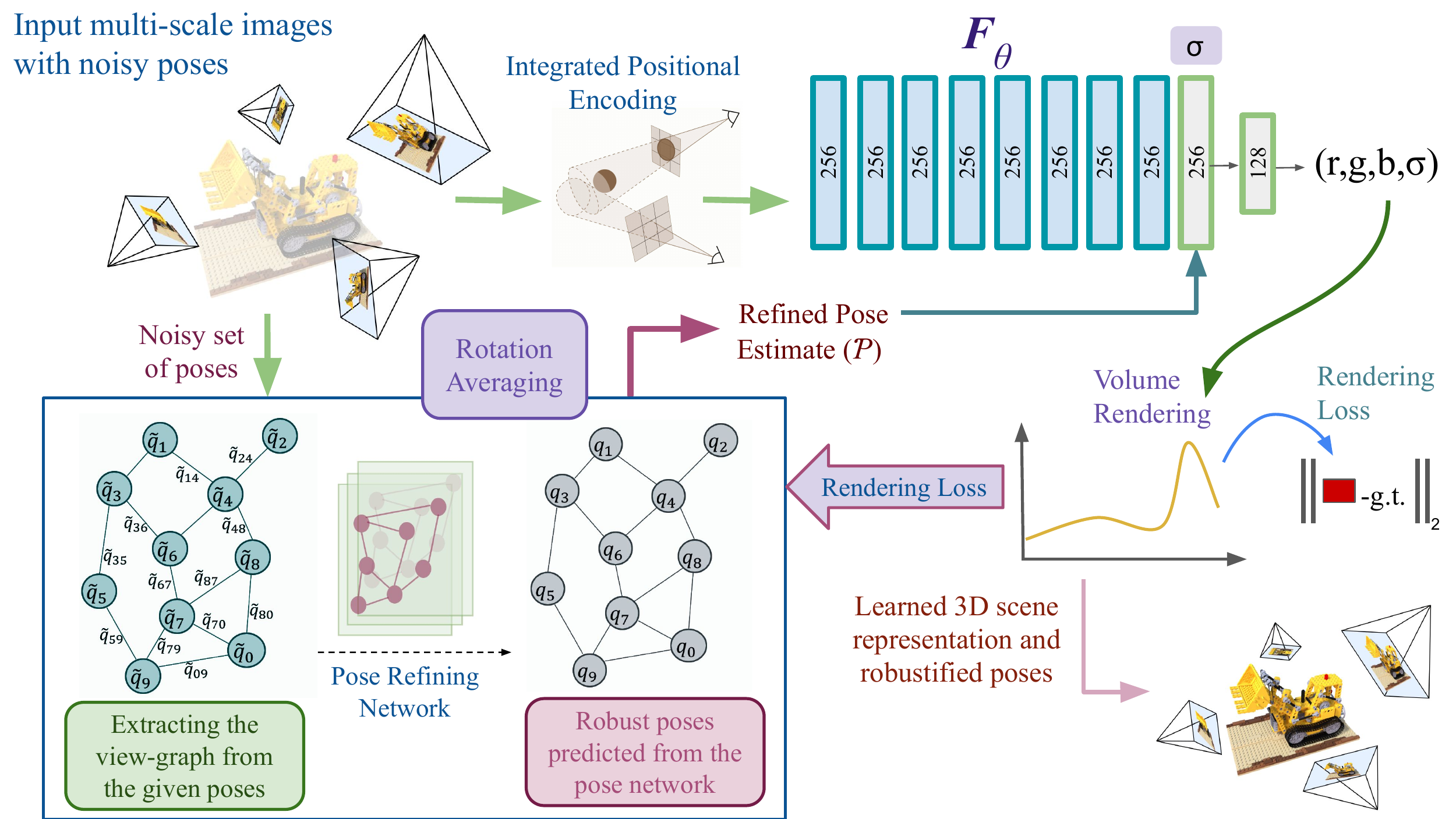}
    \caption{\footnotesize Our method jointly solve poses and learn the multi-scale object representation. The input consists of multi-scale image set and noisy set of pose. The pipeline consists of a pose-refining network to recover robust pose and estimate the IPE (Integrated Positional Encoding) by casting well-posed conical frustums through the pixels. Later, those are fed to the MLP network for learning the object 3D representation. $\mathcal{P}$ denotes set of pose.
    }
    \label{fig:pipeline}
\end{figure*}

Let $\mathcal{I}$ ~be the set of multi-view images taken at different distances from the object (see top left: Fig.\ref{fig:pipeline}). We aim to simultaneously update the MLP parameterized multi-scale representation network ($\theta$) and set of camera poses $\mathcal{P}$, given initial set of noisy estimated pose $\Tilde{\mathcal{P}}$. In probabilistic term, we can formulate it as
\begin{equation}
\label{eq:prob}
    \theta, \mathcal{P} \sim \Phi({\theta},\mathcal{P}|\mathcal{I},\Tilde{\mathcal{P}})
\end{equation}
The above formulation can further be simplified based on the assumption that scene is rigid, we can optimize for the camera pose without \emph{explicit} notion of 3d object points in the scene space. So, we simplified the Eq.\eqref{eq:prob} as follows:
\begin{equation}
    \label{eq:prob_independence}
    \Phi({\theta},\mathcal{P}|\mathcal{I},\Tilde{\mathcal{P}}) = \overbrace{\Phi(\theta|\mathcal{I}, \mathcal{P})}^{\textrm{Multiscale MLP}} ~\cdot \overbrace{\Phi(\mathcal{P}|\mathcal{I}, \Tilde{\mathcal{P}})}^{\textrm{Motion averaging}}
\end{equation}

\noindent
\textbf{Graph Neural Networks for MRA.}
Assume a directed view-graph $\mathcal{G} = (\mathcal{V, E})$ (see Fig.\ref{fig:pipeline} bottom left). A vertex $\mathcal{V}_j \in \mathcal{V}$ in this view graph corresponds to $j^{th}$ camera absolute rotation ${R}_{j}$ and $\mathcal{E}_{ij} \in \mathcal{E}$ corresponds to the relative orientation $\Tilde{R}_{ij}$ between view $i ~\text{and} ~j$ (in Fig.\ref{fig:pipeline} represented in quaternions). For our problem, relative orientations which can be noisy are used for initializing the graph. We aim to recover accurate absolute pose ${R}_{j}$ and jointly model the object representation. Conventionally, in the presence of noise, the camera motion is obtained by solving the following optimization problem 
to satisfy compatibility criteria.
\begin{equation}
\label{eq:rotation_avg_classic}
     \underset{\{{R}_{j}\} }{\text{argmin}} \sum_{\mathcal{E}_{ij} \in \mathcal{E}} \rho\Big( d(\Tilde{R}_{ij}, {R}_j {R}_i^{-1}) \Big)
\end{equation}

where, $d(.)$ denotes a suitable metric on $SO(3)$ and $\rho(.)$ is the robust loss function defined over that metric.
 Minimizing this cost function $\rho(.)$ in Eq.\eqref{eq:rotation_avg_classic} using conventional method may not be apt for several types of noise distribution observed in the real-world multi-view images. Therefore, we adhere to learn the noise distribution from the input data at train time and infer the noisy pattern to robustly predict absolute rotation. We pre-train graph neural network in a supervised setting to learn the mapping $f$ that takes noisy relative rotation $\Tilde{R}_{ij}$ and predict absolute rotations \textit{i.e.}, $\{R_j^{f}\}:=f(\Tilde{R_{ij}}; \Theta)$, where $\Theta$ is the network parameters. We train to minimize the discrepancy between ground-truth relative rotation $R_{ij} = R_j R_i^{-1}$ and estimated relative rotations $R_{ij}^{f} = R_j^{f} R_i^{-1}$ and add an extra regularizer to further learn one-to-one absolute rotation mapping.
\begin{equation}
\label{eq:RobustMRA}
     \underset{\Theta} {\text{argmin}} \sum_{\mathcal{G} \in \mathcal{D}} \sum_{\mathcal{E}_{ij} \in \mathcal{E}}  d({R}_{ij}^{f}, R_{ij}) + \beta \sum_{\mathcal{V}_{j} \in \mathcal{V}} d(R_{j}^{f}, R_j)
\end{equation}

We fix the reference rotation to be $\mathbf{I}_{3 \times 3}$ identity matrix. The mapping $f$ can now be  optimized accurately \cite{purkait2020neurora}\footnote{For more implementation details and view-graph initialization refer supplementary.}. 
Thus, our overall loss solves for accurate poses and object representation via a joint cost function. Concretely, we combine Eq.\eqref{eq:RobustMRA} ($\mathcal{
L}_{mra}$) with the squared error between the true ${C(\mathbf{r})}$ and predicted $\hat{C}(\mathbf{r}) $ pixel colors ($\mathcal{L}_{rgb}$).
\begin{equation}
\label{eq:loss_combined}
\begin{aligned}
   \mathcal{L} & = \overbrace{\sum_{\mathbf{r} \in \mathcal{R}}\|{C(\mathbf{r})} - \hat{C}(\mathbf{r})\|_2^2}^{\mathcal{L}_{rgb}} + \overbrace{\sum_{\mathcal{E}_{ij} \in \mathcal{E}}  d_{Q}({q}_{ij}^{f}, q_{ij}) + \beta \sum_{\mathcal{V}_{j} \in \mathcal{V}} d_{Q}(q_{j}^{f}, q_j)}^{\mathcal{L}_{mra}}
\end{aligned}
\end{equation}

where, $\textit{d}\textsubscript{Q} = \text{min}\{||\mathbf{p} - \mathbf{q}||_2, ||\mathbf{p} + \mathbf{q}||_2\}$ measures distance between two quaternion say $\mathbf{p}, \mathbf{q}$. Here, $\beta$ is a scalar constant. $q_{ij}$'$\text{s}$ symbolizes corresponding quaternion representation of the rotation matrix defined in Eq.\eqref{eq:RobustMRA}. $\mathcal{V}$ denotes the vertex set of the view graph corresponding to the scene being optimized and  
$\mathcal{E}$ denotes the corresponding edge set.

\subsubsection{Joint Optimization of Pose and Multi-Scale Image Rendering}


Denoting the parameters of the MLP rendering network as $\theta$ and the parameters of pose network as $\Theta$, the complete optimization objective is to search for parameters $\theta$ and $\Theta$ jointly such that loss $\mathcal{L}$ defined in Eq.(\ref{eq:loss_combined}) is minimized.
Using gradient based optimization for this search process requires calculating $\nabla_{\theta} \mathcal{L}$ and $\nabla_{\Theta} \mathcal{L}$. As $\mathcal{L}_{mra}$ is independent of rendering
network, we have $\nabla_{\theta} \mathcal{L} = \nabla_{\theta} \mathcal{L}_{rgb}$. This appears to be similar as previous optimization landscape for the rendering network, but here the poses would be changing continuously resulting in different numeric value of the gradient, making the optimization difficult to converge. Now, for the pose network $\nabla_{\Theta}\mathcal{L}$ will have 2 terms: $\nabla_{\Theta}\mathcal{L}_{rgb}$ and $\nabla_{\Theta}\mathcal{L}_{mra}$. The second term is easy to handle given the pose network is able to solve the rotations as shown in \cite{purkait2020neurora}. The first term is something that would entangle the search process for $\theta$ and $\Theta$.
For better understanding, let's assume the loss due to predicted color as
$\Phi(\theta, \mathcal{\gamma(P)})$, where $\mathcal{P}$ (with slight abuse of notation) denotes the poses having rotations predicted by the pose network, $\gamma$ denotes the positional encoding\cite{rahaman2019spectral}, then the gradient of $\Phi(\theta, \mathcal{\gamma(P)})$ w.r.t the pose network parameters $\Theta$  can be computed using backpropagation as:
\begin{equation}
    \nabla_{\Theta}\mathcal{L}_{rgb} = 
   \frac{\partial \Phi(\theta, \mathcal{\gamma(P)})}{\partial \Theta} = \frac{\partial \Phi(\theta, \mathcal{\gamma(P)})}{\partial \mathcal{\gamma(P)}} \frac{\partial \gamma(\mathcal{P})}{\partial \mathcal{P}} \frac{\partial \mathcal{P}}{\partial \Theta}
   \label{eq:joint_op_grad}
\end{equation}
Differentiating this $\gamma$ function might result in updates favourable to higher frequencies ($k$) as pointed out previously in \cite{lin2021barf}, therefore we modify this $\gamma$ function further to:
\begin{equation}
    \gamma^*(x,k) = e^{g(k)}\gamma(x)
\end{equation}
where $g(k)=\min(\frac{t-k}{b},0)$, $t$ is annealed from $0$ to maximum number of modes and $b$ is a scalar constant.
The term $\nabla_{\Theta}\mathcal{L}_{rgb}$ shown in Eq.(\ref{eq:joint_op_grad}) results in correlated updates on MLP network and pose network parameters and can result in a highly non-convex optimization.
To make optimization stable, we use the following weighted loss function:
\begin{equation}
\label{eq:weighted}
\begin{aligned}
   \mathcal{L} = \lambda \mathcal{L}_{mra} + (1-\lambda) \mathcal{L}_{rgb}
\end{aligned}
\end{equation}
where $\lambda$ is a scalar constant. Fig.(\ref{fig:pipeline}) provides the overall flow-diagram of our approach.

\smallskip
\noindent
\textbf{Optimization Strategy}. 
We begin with a disjoint optimization scheme for poses and structure, fixing $\lambda=1$ fixed for some initial number of epochs. For this case, Eq.\eqref{eq:prob_independence} depicts the modified formulation of the problem statement. 
After the initial optimization of both the networks via biased weighting strategy,  $\lambda$ is annealed by using an exponential decay, \textit{i.e.}, $\lambda=\lambda_{0}e^{-kt}$ where $\lambda_{0}=1$. This annealing goes till $\lambda=0.5$ and then we fix it at $0.5$ for the remaining optimization process. 

\vspace{-3.0mm}
\section{Experimental Setup, Results and Ablations}
\label{sec:result}
\vspace{-2.0mm}
Our approach requires optimization of two networks \textit{(i)} Graph Neural Network (GNN) for robust rotation averaging optimization based on message passing strategy \cite{gilmer2017neural, purkait2020neurora}, \textit{(ii)} Multi-layer Perceptron (MLP) network optimization for neural multi-scale scene representation. Our GNN architecture for pose optimization is inspired from Purkait \textit{et al.} \cite{purkait2020neurora} FineNet, whereas the MLP based rendering network is similar to Mip-NeRF \cite{barron2021mipnerf}. 
Implementation details for reproducibility and hyperparameter values are provided in the supplementary. Also, pre-training scheme of our GNN, extensive evaluation on pose graphs and robustness to noisy correspondences are provided in the supplementary.

\smallskip
\noindent
\textbf{Baselines.} We compared our method's performance with the existing methods that either resolves multi-scale or pose problems with NeRF \cite{mildenhall2020nerf} such as BARF\cite{lin2021barf}, NeRF--\cite{wang2021nerf}, and Mip-NeRF\cite{barron2021mipnerf}. To further show the effectiveness of our method, we define new baselines. Our newly defined baseline A (Base A) combines BARF and Mip-NeRF loss. This is done by making the poses input to the Mip-NeRF trainable and updating the positional encoding scheme similar to BARF. Next, we define baseline B (Base B) where we first run BARF on the multi-scale scene and then train the Mip-NeRF model with the output poses. Finally, we define baseline C (Base C) where we combine Mip-NeRF and NeRF-- by just updating NeRF-- with the Mip-NeRF positional encoding scheme. Refer supplementary for further reasoning behind using these baselines.


\subsection{Test sets and Results}
\label{sec:mutlib}
To compare our method with the baselines, we used the Blender dataset provided by the authors of NeRF \cite{mildenhall2020nerf}\footnote{CC-BY-3.0 license.} and its multi-scale version provided by Barron \textit{et al.} \cite{barron2021mipnerf}. It consists of single object scenes comprising synthetic objects and corresponding ground truth poses, with each scene consisting of $M=100$ images with $800\times800$ resolution.
To simulate the effect of pose estimation errors in real-world datasets, we add noise to the ground truth poses. More details regarding datasets are provided in the following subsections. Furthermore, we also test our approach on real-world dataset\cite{Knapitsch2017}, where COLMAP poses are treated as G.T., and show improvements over these roughly accurate poses in the supplementary. 

\smallskip
\noindent
\textbf{Multi-Scaled Images of Object.} We first study the multi-scaled version of the Blender dataset proposed in the Mip-NeRF. It consists of 400 image scenes generated by scaling every image in the original Blender dataset to 4 different resolutions. These different resolution images are synthesized by concatenation of actual resolution images with downsampled images by a factor of 2, 4, and 8. This scale can also be interpreted as the distance of the object from the camera. Therefore, it resembles the real-world datasets much more closely than the original Blender dataset, which contains all the images at the same image resolution and nearly similar distances. The ground truth extrinsic poses are the same as the original dataset, but the camera intrinsics are changed according to the image resolution. We perturb the poses for every scene by first sampling the noise from a normal distribution $\delta$\textbf{p} $\sim \mathcal{N}(\textbf{0},1e^{-1}\textbf{I})$, and adding the noise to rotation in its axis-angle form. It is then converted to the rotation matrix representation and multiplied with the ground truth poses, disturbing their orientations.
We did this purposely to make the dataset resemble real-world settings closely, thus making it challenging to learn the multi-scale scene representation.

\begin{table}[t]
\centering
\resizebox{\textwidth}{!}
{
    \begin{tabular}{ccccccccccccccccc}
        \multicolumn{1}{c|}{}                                        & \multicolumn{2}{c|}{Lego}                         &  \multicolumn{2}{c|}{Ship}     & \multicolumn{2}{c|}{Drums}   & \multicolumn{2}{c|}{Mic}
        & \multicolumn{2}{c|}{Chair}
        & \multicolumn{2}{c|}{Ficus} & \multicolumn{2}{c|}{Materials}  &
        \multicolumn{2}{c}{Hotdog} \\
        
        \multicolumn{1}{c|}{}                                                                          & \multicolumn{1}{c}{{\fontsize{6.5}{4}\selectfont PSNR$\uparrow$}} & \multicolumn{1}{c|}{{\fontsize{6.5}{4}\selectfont LPIPS$\downarrow$}} &  \multicolumn{1}{c}{{\fontsize{6.5}{4}\selectfont PSNR$\uparrow$}} & \multicolumn{1}{c|}{{\fontsize{6.5}{4}\selectfont LPIPS$\downarrow$}} &                     \multicolumn{1}{c}{{\fontsize{6.5}{4}\selectfont PSNR$\uparrow$}} & \multicolumn{1}{c|}{{\fontsize{6.5}{4}\selectfont LPIPS$\downarrow$}} & \multicolumn{1}{c}{{\fontsize{6.5}{4}\selectfont PSNR$\uparrow$}} & \multicolumn{1}{c|}{{\fontsize{6.5}{4}\selectfont LPIPS$\downarrow$}} & 
        \multicolumn{1}{c}{{\fontsize{6.5}{4}\selectfont PSNR$\uparrow$}} & \multicolumn{1}{c|}{{\fontsize{6.5}{4}\selectfont LPIPS$\downarrow$}} & 
        \multicolumn{1}{c}{{\fontsize{6.5}{4}\selectfont PSNR$\uparrow$}} & \multicolumn{1}{c|}{{\fontsize{6.5}{4}\selectfont LPIPS$\downarrow$}}
        & 
        \multicolumn{1}{c}{{\fontsize{6.5}{4}\selectfont PSNR$\uparrow$}} & \multicolumn{1}{c|}{{\fontsize{6.5}{4}\selectfont LPIPS$\downarrow$}}
        &
        \multicolumn{1}{c}{{\fontsize{6.5}{4}\selectfont PSNR$\uparrow$}} & \multicolumn{1}{c}{{\fontsize{6.5}{4}\selectfont LPIPS$\downarrow$}}
        
        \\ \hline
        \multicolumn{1}{c|}{\begin{tabular}[c]{@{}c@{}}Mip-NeRF\end{tabular}} & 21.52   & \multicolumn{1}{c|}{0.06} &  \cellcolor[HTML]{ffff80}24.54 &  \multicolumn{1}{c|}{ \cellcolor[HTML]{ffff80}0.07} & 13.34   & \multicolumn{1}{c|}{0.075}    &  \cellcolor[HTML]{ffff80}24.71  & \multicolumn{1}{c|}{ \cellcolor[HTML]{ffff80}0.05} & \cellcolor[HTML]{ffff80}29.1  &  \multicolumn{1}{c|}{\cellcolor[HTML]{ffff80}0.049}  & 22.47  &  \multicolumn{1}{c|}{0.055} & 19.7  &  \multicolumn{1}{c|}{0.089}& \cellcolor[HTML]{ffff80}27.09  &  \multicolumn{1}{c}{\cellcolor[HTML]{ffff80}0.053}        \\ 
        \multicolumn{1}{c|}{\begin{tabular}[c|]{@{}c@{}}BARF\end{tabular}}          & 10.88   & \multicolumn{1}{c|}{0.55} & 8.81 &  \multicolumn{1}{c|}{0.74}  & 11.56  &  \multicolumn{1}{c|}{0.76}  & 12.35  &  \multicolumn{1}{c|}{0.57}  & 14.35  &  \multicolumn{1}{c|}{0.47}  & 11.88  &  \multicolumn{1}{c|}{0.65}  & 12.28  &  \multicolumn{1}{c|}{0.61}
        & 14.28  &  \multicolumn{1}{c}{0.46}
        \\     \hline
        
        \multicolumn{1}{c|}{\begin{tabular}[c]{@{}c@{}}Base A\end{tabular}} & 11.67   & \multicolumn{1}{c|}{0.49} &  14.28 &  \multicolumn{1}{c|}{0.28} & 13.25  &  \multicolumn{1}{c|}{0.67}    & 12.28  & \multicolumn{1}{c|}{0.41}  & 15.12  &  \multicolumn{1}{c|}{0.20}  & 12.31  &  \multicolumn{1}{c|}{0.25} & 13.31  &  \multicolumn{1}{c|}{0.42}        & 16.17  &  \multicolumn{1}{c}{0.39}  \\ 
        \multicolumn{1}{c|}{\begin{tabular}[c]{@{}c@{}}Base B\end{tabular}} & 12.46 &  \multicolumn{1}{c|}{0.37} & 13.43 &   \multicolumn{1}{c|}{0.31} & 11.32  &  \multicolumn{1}{c|}{0.58}    &  14.26  &  \multicolumn{1}{c|}{ 0.29} & 13.71  &  \multicolumn{1}{c|}{0.42}  & 11.56  &  \multicolumn{1}{c|}{0.52}     & 12.22  &  \multicolumn{1}{c|}{0.47}& 15.87  &  \multicolumn{1}{c}{0.42}      \\ \hline
         \multicolumn{1}{c|}{\begin{tabular}[c]{@{}c@{}}NeRF--\end{tabular}} & 16.89 &  \multicolumn{1}{c|}{0.094} &  19.89 &   \multicolumn{1}{c|}{0.118} & 15.67  &  \multicolumn{1}{c|}{0.074}    &  18.35  &  \multicolumn{1}{c|}{ 0.08}  & 20.22  &  \multicolumn{1}{c|}{0.098}  & 14.44  &  \multicolumn{1}{c|}{0.13} & 15.77  &  \multicolumn{1}{c|}{0.22} & 18.69  &  \multicolumn{1}{c}{0.20}         \\ 
        \multicolumn{1}{c|}{\begin{tabular}[c]{@{}c@{}}Base C\end{tabular}} & 18.28 &  \multicolumn{1}{c|}{0.089} &  16.32 &   \multicolumn{1}{c|}{0.22} & 17.25  &  \multicolumn{1}{c|}{0.070}    &  19.42  &  \multicolumn{1}{c|}{ 0.073}  & 18.67  &  \multicolumn{1}{c|}{0.114}  & 16.32  &  \multicolumn{1}{c|}{0.12} & 16.58  &  \multicolumn{1}{c|}{0.207} & 17.55  &  \multicolumn{1}{c}{0.223}         \\
        \multicolumn{1}{c|}{\begin{tabular}[c]{@{}c@{}}Ours\end{tabular}} & \cellcolor[HTML]{ff9090} \textbf{ 27.01}    &  \multicolumn{1}{c|}{\cellcolor[HTML]{ff9090} \textbf{0.044}} & \cellcolor[HTML]{ff9090} \textbf{ 26.59}  & \multicolumn{1}{c|}{\cellcolor[HTML]{ff9090} \textbf{ 0.067}} & \cellcolor[HTML]{ff9090} \textbf{ 26.07}   & \multicolumn{1}{c|}{\cellcolor[HTML]{ff9090} \textbf{0.043}} & \cellcolor[HTML]{ff9090} \textbf{32.8}  &   \multicolumn{1}{c|}{\cellcolor[HTML]{ff9090} \textbf{0.008}} &  \cellcolor[HTML]{ff9090}\textbf{35.23}  &   \multicolumn{1}{c|}{\cellcolor[HTML]{ff9090} \textbf{0.031}} &  \cellcolor[HTML]{ff9090}\textbf{29.28}  &   \multicolumn{1}{c|}{\cellcolor[HTML]{ff9090}\textbf{0.032}} & \cellcolor[HTML]{ff9090}\textbf{24.8}  &  \multicolumn{1}{c|}{\cellcolor[HTML]{ff9090}\textbf{0.061}}
        & \cellcolor[HTML]{ff9090}\textbf{32.5}  &  \multicolumn{1}{c}{\cellcolor[HTML]{ff9090}\textbf{0.028}}
    \end{tabular}
}
\caption{\footnotesize Performance comparison with other competing approaches. We used widely used PSNR and LPIPS performance metric to document the results. Clearly, our method supersede the results of BARF\cite{lin2021barf}, Mip-NeRF\cite{barron2021mipnerf} and NeRF--\cite{wang2021nerf} on the multi-scale blender synthetic dataset proposed by \cite{barron2021mipnerf}. The details of other baselines are given in Sec. \S \ref{sec:result}. For this experiment, we synthetically perturbed the poses. The results suggest that our approach can favorably deal with multi-scale and camera pose problem.
}\label{table:1}
\end{table}

\begin{figure}[t]
    \centering
    \includegraphics[scale=0.41]{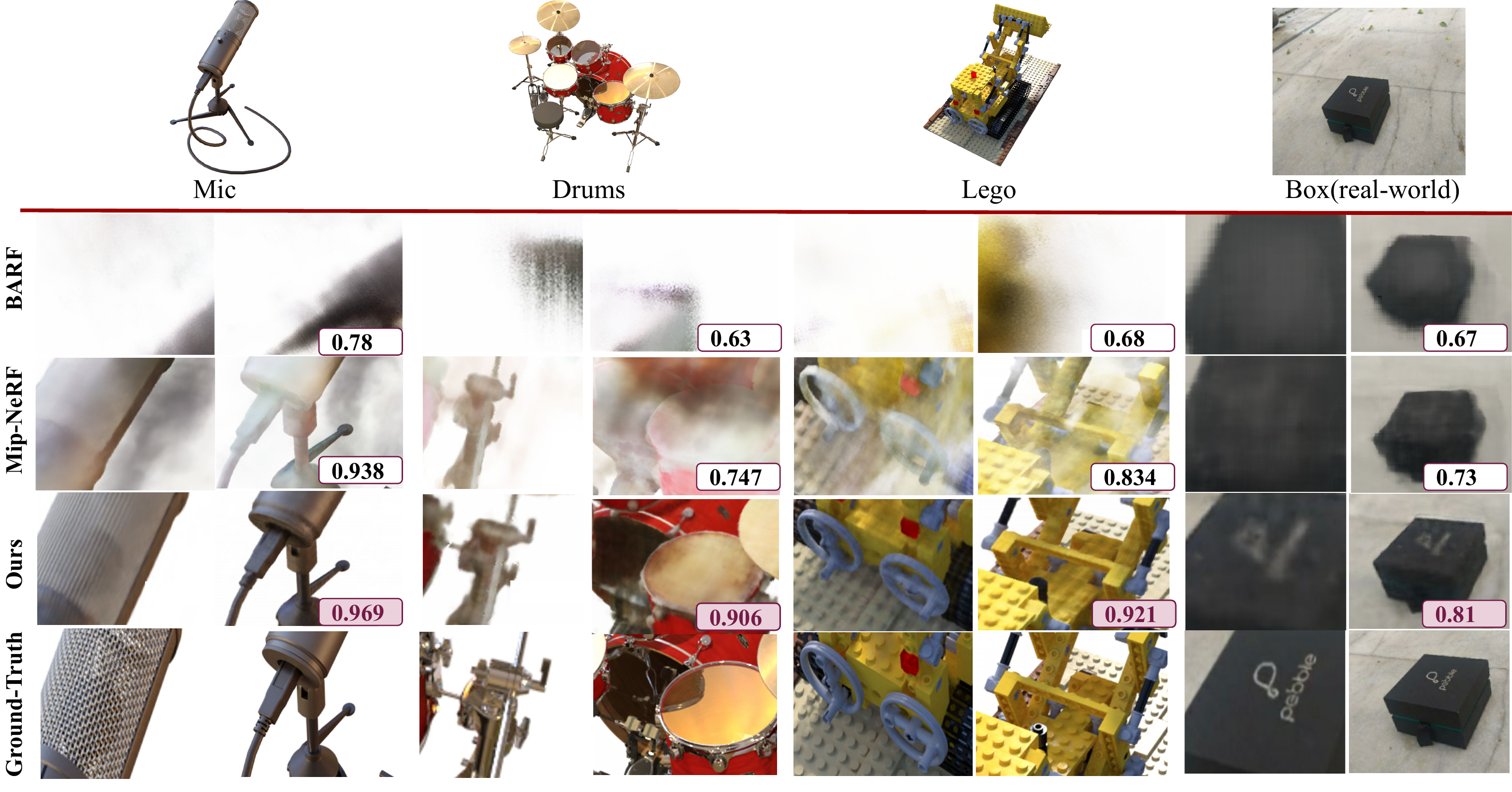}
    \caption{\footnotesize Qualitative Comparison of our method, Mip-NeRF\cite{barron2021mipnerf} and BARF\cite{lin2021barf} on the multi-scale of blender dataset with synthetically added pose errors. We have visualized the synthesized images from all the approaches on 3 scenes namely Mic, Drums, Lego and our real sequence. Our method clearly provide better synthesized images.
    }
    \label{fig:vis}
\end{figure}

Table (\ref{table:1}) provide the results with multi-scale images and pose error as input. The results are compared using the popular PSNR and LPIPS metric averaged across all the four resolution images. 
The results show that our method can jointly solve the multi-scale and pose problems with NeRF and gives results better than other baseline approaches.
Fig.(\ref{fig:vis}) provide the qualitative result comparison for the same.

\smallskip
\noindent
\textbf{Tanks and Temples.} Tanks and Temples is a well-known challenging dataset containing real-world scenes [9].
It consists of images showing large scale scenes to simulating the realistic conditions and
largely used for evaluating 3D reconstruction methods. Further, this dataset can be very
useful for testing unconstrained view-synthesis methods. Accordingly, we used couple of sequence to test our method’s performance. 
Specifically, we used ``truck'' and ``tank'' sequence, which consists of image set containing a 360$^{\circ}$ view of the subject captured 
freely at a varying distance from the object. Since there are no ground-truth poses provided by the dataset, we used COLMAP\cite{schoenberger2016sfm} to 
estimate the initial poses and feed them to our network. Table (\ref{table:col}) shows the comparison between our method and Mip-NeRF\cite{barron2021mipnerf} for these two sequence. Clearly, our method efficiently optimizes over the poses esimated by the COLMAP and is able to generate better image renderings when compared Mip-NeRF+COLMAP setting. For more details regarding initial camera estimation, 
evaluation details on this dataset and experimental observations, please refer supplementary.

\smallskip
\noindent
\textbf{Randomly Captured Black-Box Sequence.} In Fig.(\ref{fig:vis}), we also introduced a new sequence, containing randomly captured images of a black box, imitating a general purpose multi-image acquisition \textit{i.e.}, multi-scale and with non-smooth pose trajectory. Again, we use COLMAP to estimate the initial poses. From the results, it can be concluded that our approach clearly outperforms all the existing methods in this realistic scenario due to its robust pipeline jointly estimating scene and structure. Refer supplementary for more details regarding the images and the COLMAP estimated cameras for this sequence.

\begin{table}[t]
\centering
\scriptsize
\resizebox{0.6\columnwidth}{!}
{
    \begin{tabular}{ccccccc}
        \multicolumn{1}{c|}{}& \multicolumn{3}{c|}{MipNeRF} &  \multicolumn{3}{c}{Ours}\\ 
        \multicolumn{1}{c|}{} & \multicolumn{1}{c}{{\fontsize{6.5}{4}\selectfont PSNR$\uparrow$}} & \multicolumn{1}{c}{{\fontsize{6.5}{4}\selectfont LPIPS$\downarrow$}} & 
        \multicolumn{1}{c|}{{\fontsize{6.5}{4}\selectfont SSIM$\uparrow$}} &
        \multicolumn{1}{c}{{\fontsize{6.5}{4}\selectfont PSNR$\uparrow$}} & \multicolumn{1}{c}{{\fontsize{6.5}{4}\selectfont LPIPS$\downarrow$}} &                     \multicolumn{1}{c}{{\fontsize{6.5}{4}\selectfont SSIM$\uparrow$}} 
        \\ \hline
         \multicolumn{1}{c|}{\begin{tabular}[c]{@{}c@{}}Truck\end{tabular}} &  23.1 &  0.308 & \multicolumn{1}{c|}{ 0.812} &  \textbf{24.7} &  \textbf{0.296} & \multicolumn{1}{c}{ \textbf{0.828}}         \\
         \hline
        \multicolumn{1}{c|}{\begin{tabular}[c]{@{}c@{}}Tank\end{tabular}} &  24.8 &  0.313 & \multicolumn{1}{c|}{ 0.823} &  \textbf{26.6} &  \textbf{0.302} & \multicolumn{1}{c}{ \textbf{0.851}}         \\
    \end{tabular}
}
\caption{\footnotesize Performance comparison of our approach and Mip-NeRF\cite{barron2021mipnerf} on the Truck and Tank sequence \cite{Knapitsch2017}.  The result shows that our method performs better as compared to Mip-NeRF on both the freely moving sequences demonstrating our method's advantage. {It can be inferred from the above statistics that just relying on COLMAP poses for solving image based rendering on unconstrained sequence can demonstrably give inferior results. On the contrary, our approach can handle bad camera poses and provide favorable novel view rendering.} 
}\label{table:col}
\end{table}

\subsection{Ablations}\label{sec:ablations}

\noindent
\textbf{(a) Same Scale Images with Pose Error.} 
Similar to the multi-scale case setup as described in \S Sec.\ref{sec:mutlib}, we perturb the pose estimates of the Blender dataset, which contains object multi-view images captured from the same distance. Table (\ref{table:3}) shows the PSNR, LPIPS, and SSIM results for this setting corresponding to four scenes in the dataset namely \textit{Lego}, \textit{ship}, \textit{drums} and \textit{mic}. The performance confirms that our method more often than not supersedes the baselines results with similar scale images. This is expected as BARF \cite{lin2021barf} was designed to handle pose for images taken at same distant from the object.


\begin{table}[!htb]
\scriptsize
\centering
\resizebox{\columnwidth}{!}
{
\begin{tabular}{ccccccccccccc}
        \multicolumn{1}{c|}{}                                        & \multicolumn{3}{c|}{Lego}                         &  \multicolumn{3}{c|}{Ship}     & \multicolumn{3}{c|}{Drums}   & \multicolumn{3}{c}{Mic}                     \\ 
        \multicolumn{1}{c|}{}                                                                          & \multicolumn{1}{c}{{\fontsize{6.5}{4}\selectfont PSNR}} & \multicolumn{1}{c}{{\fontsize{6.5}{4}\selectfont LPIPS}} & \multicolumn{1}{c|}{{\fontsize{6.5}{4}\selectfont SSIM}}                                                                                                                                           & \multicolumn{1}{c}{{\fontsize{6.5}{4}\selectfont PSNR}} & \multicolumn{1}{c}{{\fontsize{6.5}{4}\selectfont LPIPS}} & \multicolumn{1}{c|}{{\fontsize{6.5}{4}\selectfont SSIM}}                           & \multicolumn{1}{c}{{\fontsize{6.5}{4}\selectfont PSNR}} & \multicolumn{1}{c}{{\fontsize{6.5}{4}\selectfont LPIPS}} & \multicolumn{1}{c|}{{\fontsize{6.5}{4}\selectfont SSIM}}    & \multicolumn{1}{c}{{\fontsize{6.5}{4}\selectfont PSNR}} & \multicolumn{1}{c}{{\fontsize{6.5}{4}\selectfont LPIPS}} & \multicolumn{1}{c}{{\fontsize{6.5}{4}\selectfont SSIM}}    
        
        \\ \hline
        \multicolumn{1}{c|}{\begin{tabular}[c]{@{}c@{}}Mip-NeRF\end{tabular}} & 17.90 & 0.089   & \multicolumn{1}{c|}{0.82} & 22.90 & 0.107 & \multicolumn{1}{c|}{0.71} & 14.07  & 0.11 & \multicolumn{1}{c|}{0.799}    & 21.90  & 0.064 & \multicolumn{1}{c}{0.93}    \\ 
        \multicolumn{1}{c|}{\begin{tabular}[c|]{@{}c@{}}BARF\end{tabular}}          &  \cellcolor[HTML]{ff9090} 27.61  & \cellcolor[HTML]{ffff80} 0.05 & \multicolumn{1}{c|}{\cellcolor[HTML]{ff9090} 0.92} &  \cellcolor[HTML]{ff9090} 26.18 & \cellcolor[HTML]{ff9090} 0.121 & \multicolumn{1}{c|}{\cellcolor[HTML]{ffff80} 0.74}  & \cellcolor[HTML]{ffff80}23.68  & \cellcolor[HTML]{ffff80} 0.095 & \multicolumn{1}{c|}{\cellcolor[HTML]{ffff80} 0.88}  & \cellcolor[HTML]{ffff80} 27.03  & \cellcolor[HTML]{ffff80} 0.06 & \multicolumn{1}{c}{\cellcolor[HTML]{ffff80}0.96}                   \\     \hline
       
        \multicolumn{1}{c|}{\begin{tabular}[c]{@{}c@{}}RM-NeRF(ours)\end{tabular}} & \cellcolor[HTML]{ffff80} 27.10 & \cellcolor[HTML]{ff9090} 0.048   & \multicolumn{1}{c|}{ \cellcolor[HTML]{ff9090} 0.92} & \cellcolor[HTML]{ffff80} 25.45 & \cellcolor[HTML]{ff9090} 0.0690 & \multicolumn{1}{c|}{\cellcolor[HTML]{ffff80} 0.735} &  \cellcolor[HTML]{ff9090} 24.98  & \cellcolor[HTML]{ff9090} 0.072 & \multicolumn{1}{c|}{\cellcolor[HTML]{ff9090}  0.907} &  \cellcolor[HTML]{ff9090} 30.03  & \cellcolor[HTML]{ff9090} 0.027 & \multicolumn{1}{c}{\cellcolor[HTML]{ff9090} 0.963} \\ 
        
    \end{tabular}
    }
\caption{\footnotesize  PSNR, LPIPS and SSIM comparison of our method with BARF\cite{lin2021barf} and Mip-NeRF\cite{barron2021mipnerf} on Blender dataset\cite{mildenhall2020nerf} with synthetically introduced pose errors.}
\label{table:3}
\end{table}

\begin{table}[!htb]
\scriptsize
\centering
\resizebox{\columnwidth}{!}
{
\begin{tabular}{ccccccccccccc}
        \multicolumn{1}{c|}{}                                        & \multicolumn{3}{c|}{Lego}                         &  \multicolumn{3}{c|}{Ship}     & \multicolumn{3}{c|}{Drums}   & \multicolumn{3}{c}{Mic}                      \\ 
        \multicolumn{1}{c|}{}                                                                          & \multicolumn{1}{c}{{\fontsize{6.5}{4}\selectfont PSNR}} & \multicolumn{1}{c}{{\fontsize{6.5}{4}\selectfont LPIPS}} & \multicolumn{1}{c|}{{\fontsize{6.5}{4}\selectfont SSIM}}                                                                                                                                           & \multicolumn{1}{c}{{\fontsize{6.5}{4}\selectfont PSNR}} & \multicolumn{1}{c}{{\fontsize{6.5}{4}\selectfont LPIPS}} & \multicolumn{1}{c|}{{\fontsize{6.5}{4}\selectfont SSIM}}                           & \multicolumn{1}{c}{{\fontsize{6.5}{4}\selectfont PSNR}} & \multicolumn{1}{c}{{\fontsize{6.5}{4}\selectfont LPIPS}} & \multicolumn{1}{c|}{{\fontsize{6.5}{4}\selectfont SSIM}}    & \multicolumn{1}{c}{{\fontsize{6.5}{4}\selectfont PSNR}} & \multicolumn{1}{c}{{\fontsize{6.5}{4}\selectfont LPIPS}} & \multicolumn{1}{c}{{\fontsize{6.5}{4}\selectfont SSIM}}    
        
        \\ \hline
        \multicolumn{1}{c|}{\begin{tabular}[c]{@{}c@{}}RM-NeRF(ours$^\dagger$)\end{tabular}} & 22.20 & 0.067   & \multicolumn{1}{c|}{0.87} & 23.34 & 0.071 & \multicolumn{1}{c|}{0.71} & 15.07  & 0.079 & \multicolumn{1}{c|}{0.789}    & 22.60  & 0.049 & \multicolumn{1}{c}{0.92}    \\ \hline
       
        \multicolumn{1}{c|}{\begin{tabular}[c]{@{}c@{}}RM-NeRF(ours)\end{tabular}} &  27.01 &  0.044   & \multicolumn{1}{c|}{ 0.92} &  26.59 &  0.067 & \multicolumn{1}{c|}{ 0.74} &  26.07  &  0.043 & \multicolumn{1}{c|}{ 0.92} &  32.8  &  0.008 & \multicolumn{1}{c}{ 0.97} \\ 
    \end{tabular}
    }
\caption{\footnotesize Comparison of our proposed optimization method (ours) with its variant (ours$^\dagger$) where we fix $\lambda=0.5$ on Mulit-scale Blender dataset with introduced pose errors.}
\label{table:4}
\end{table}
\noindent
\textbf{(b) Unbiased Optimization of Eq. \eqref{eq:weighted} ($\boldsymbol{\lambda=0.5}$)}. 
To better understand the behaviour of our joint optimization and utility of our annealing strategy, we conducted this experiment. By setting $\lambda=0.5$ in Eq.\eqref{eq:weighted}, we assign equal weight to color rendering cost ($\mathcal{L}_{rgb}$)  and robust MRA loss ($\mathcal{L}_{mra}$) loss during optimization. 
%
Table (\ref{table:4}) shows a comparison of this method with our optimization method on the 4 scenes of the multi-scale blender dataset with synthetic noise. The statistics indicate that utilizing the rigid scene prior during optimization helps. Furthermore, in supplementary, we perform another ablation where we compare all the methods on the Multi-scale Blender dataset using the available ground truth poses.
\vspace{-2.0mm}
\section{Conclusion and Future Direction}
We introduced an approach that enhances the use of neural radiance fields representation to general daily acquired multi-view images, where multi-scale images and camera pose errors are inevitable. By unifying the concepts from multi-view geometry in computer vision, multi-scale NeRF, and graph neural networks, we propose a method that can robustly solve multi-scale image rendering issues in continuous volume rendering. Of course, the proposed method is not a perfect solution to the problem; however, it suggests an important area for research that could enable continuous neural volume rendering to daily acquired multi-view images. A straightforward future direction is to extend the proposed approach for jointly estimating the camera intrinsics and the extrinsic for multi-scale neural scene representation.

\bibliography{egbib}

\end{document}